\documentclass{article}

\usepackage{microtype}
\usepackage{graphicx}
\usepackage{subfigure}
\usepackage{booktabs} 
\usepackage{hyperref}

\usepackage{tikz}
\usetikzlibrary{matrix,positioning,shadows.blur,decorations.pathreplacing,backgrounds,fit,decorations.markings}
\usepackage{calc}
\usepackage{fp}
\usepackage{xparse}
\usepackage{float}

\usepackage[accepted]{icml2020}

\usepackage{amssymb}
\usepackage{calc}
\usepackage{color, colortbl}
\usepackage{fp}
\usepackage{latexsym}
\usepackage{listings}
\usepackage{pgfpages}

\usepackage{tikz}
\usepackage{times}
\usepackage{varwidth}
\usepackage{xparse}
\usepackage{xspace}
\usetikzlibrary{matrix,positioning,shadows.blur,decorations.pathreplacing,backgrounds,fit,shapes.geometric}

\ExplSyntaxOn
\DeclareExpandableDocumentCommand{\eval}{m}{\int_eval:n {#1}}
\ExplSyntaxOff

\usepackage{microtype}

\begin{document}

\twocolumn[
\icmltitle{Graph-Convolutional Autoencoder Ensembles for the Humanities, Illustrated with a Study of the American Slave Trade}


\icmlsetsymbol{equal}{*}

\begin{icmlauthorlist}
\icmlauthor{Tom Lippincott}{tom}
\end{icmlauthorlist}

\icmlaffiliation{tom}{Center for Language and Speech Processing, Johns Hopkins University, Baltimore, Maryland, USA}

\icmlcorrespondingauthor{Tom Lippincott}{tom@cs.jhu.edu}

\icmlkeywords{Machine Learning, Humanities}

\vskip 0.3in
]

\printAffiliationsAndNotice{}  

\begin{abstract}
  We introduce \emph{Starcoder}, a graph-aware autoencoder ensemble framework, with associated formalisms and tooling, designed to facilitate deep learning for scholarship in the humanities.  By composing sub-architectures to produce a model isomorphic to a humanistic domain we maintain interpretability while providing function signatures for each sub-architectural choice, allowing both traditional and computational researchers to collaborate without disrupting established practices.  We illustrate a practical application of our approach to a historical study of the American post-Atlantic slave trade, and make several specific technical contributions: a novel hybrid graph-convolutional autoencoder mechanism, batching policies for common graph topologies, and masking techniques for particular use-cases.  The effectiveness of the framework for broadening participation of diverse domains is demonstrated by a growing suite of two dozen studies, both collaborations with humanists and established tasks from machine learning literature, spanning a variety of fields and data modalities.  We make performance comparisons of several different architectural choices and conclude with an ambitious list of imminent next steps for this research.
\end{abstract}

\section{Introduction}

A major downstream use-case for machine learning is in support of \emph{knowledge workers}, highly-skilled individuals responsible for maintaining and communicating a deep understanding of a subject.  While large organizations in government and industry are avid consumers and funders of cutting edge machine learning, there is a long-standing struggle to connect with academics working in the humanities.  The barriers are manifold, but a pervasive problem is the \emph{disruption to existing scholarly methods} that, in most cases, requires a humanist to internalize computational skills and mindset before reaping benefits.  Such disription is beyond the reach of a time-constrained graduate student, and consequently, in the best of circumstances machine learning in the humanities is mostly limited to a handful of methods that lend themselves to immediate interpretation or obvious utility, such as topic modeling \cite{tm} or optical character recognition \cite{ocr}.

At the same time, machine learning researchers have made progress on several related areas: the focus on \emph{interpretability} \cite{blackboxnlp}, while often motivated by the researchers themselves, is a critical ingredient for translating results into the academic currency of the target field.  At the moment, this is an ad hoc process that requires a machine learning researcher to either be deeply familiar with the target field, or work closely with traditional scholars to translate the computational output, neither of which is typically recognized for career advancement.

In this paper we introduce \emph{Starcoder}, a framework designed to allow researchers both in machine learning and the humanities to remain focused on their field-specific goals, while giving and recieving benefits from one another via a well-defined orchestration of primary sources, formalisms, open sets of neural architectures, and exploratory interfaces.

From the perspective of a computational researcher, we make the following contributions:

\begin{itemize}
\item Automatic generation of multi-modal neural ensembles based on well-defined formalisms
\item A novel combination of autoencoding and graph-convolutional mechanisms
\item Abstractions for the major sub-architectural choices that are easy for machine learning researchers to target and reason about
\item Experiments comparing choices for several orthogonal components of the generation process
\end{itemize}

From the perspective of a traditional scholar, we make the following contributions:

\begin{itemize}
\item Low-barrier entry to modern deep learning for bespoke humanistic domains
\item Relevant examples including benefits to real-world scholarship
\item Automatic transition from machine learning to public digital humanities
\end{itemize}

While this paper focuses on the creation of an inclusive, extensible path from the humanities to machine learning, the appendix further describes the scholar- and public-facing interfaces Starcoder automatically produces as artifacts of this process, and that enable humanists to explore, annotate, and learn from their primary sources, and ultimately present insights to the public.  Screenshots with descriptions are provided, and an example of this interface generated directly from our current collaborations can be viewed at \url{https://www.comp-int-hum.org}.

\section{Representing a traditional domain}

The linchpin of our approach is a formal specification (\emph{schema}) of the traditional scholar's domain of interest as an \emph{entity-relationship model} (ERM) \cite{chen:2002} where \emph{entity-types} have potential \emph{properties} and \emph{relationships}.  This information allows Starcoder to generate an isomorphic model, and for a trained model to be explored, reused, and interacted with intuitively.

\begin{figure}[h!]
  \centering
  \lstset{basicstyle=\tiny}
  \begin{lstlisting}
{
  "@context": {
    "@vocab": "https://www.comp-int-hum.org"
  }
  "entity_types": {
    "person": ["name", "age", "job"],
    "location": ["coordinates", "photo"]
  },
  "properties": {
    "name": {"type": "text"},
    "age": {"type": "scalar"},
    "job": {"type": "categorical"},
    "coordinates": {"type": "place"},
    "photo": {"type": "image"}
  },
  "relationships": {
    "office_of": {
      "source_entity_type": "location",
      "target_entity_type": "person"
    },    
    "client_of": {
      "source_entity_type": "person",
      "target_entity_type": "person"
    }
  }
}
  \end{lstlisting}
  \caption{A simple domain is described in terms of its entities, their properties, and their relationships.}
  \label{figure:domain}
\end{figure}

ERMs are closely related to knowledge graphs \cite{kb}, relational databases \cite{relationaldatabases}, ontologies \cite{ontologies}, first-order predicate logic \cite{frege}, and a host of linguistic formalisms.  Most importantly, schemas capture the structure of a domain in a precise, flexible fashion that enables both traditional and computational studies without disrupting existing research practices.  To this end, we employ JSON-LD \cite{jsonld}, an intuitive, RDF-compatible format, as the schema format.  Figure \ref{figure:domain} shows a simple example that will be used later to illustrate model generation: while expressive and extensible, the format can be easily understood and edited by humanists, seeded heuristically from other structured formats (XML, CSV, CONLL, etc), or represented graphically.

\begin{figure}[h!]
  \centering
  \lstset{basicstyle=\tiny}
  \begin{lstlisting}
{
  "entity_type": "person",
  "id": "P1",
  "name": "Mary",
  "age": 27,
}
{
  "entity_type": "location",
  "id": "L1",
  "coordinates": {
    "latitude" : 39.29,
    "longitude" : 76.61
  },
  "photo": "www.site.com/shot.jpg",
  "office_of": ["P1", "P4"]
}
  \end{lstlisting}
  \caption{Example entities following the schema in Figure \ref{figure:domain}.}
  \label{figure:entity}
\end{figure}

Figure \ref{figure:entity} shows two entities following the domain in Figure \ref{figure:domain}: again, they are standard JSON and easily interpreted and edited by humanists.  There are two metadata properties (\emph{entity\_type} and \emph{id}), regular properties of various types, and a one-to-many relationship property, \emph{office\_of}.\footnote{Starcoder makes no distinction between relationship cardinality, though an optional field would allow data to be checked for correctness.}

\begin{figure}[h!]
  \centering
  \lstset{basicstyle=\tiny}
  \begin{lstlisting}
[
  "$.properties[?(@.type=='image')]",
  {
    "width": 32,
    "height": 32,
    "channels": 3,
    "channel_size": 8,
    "decoder": "NullDecoder"
  }
]
  \end{lstlisting}
  \caption{Example JSONPath rule setting all image properties to down-sample inputs to a particular shape and encoding.}
  \label{figure:jsonpath}
\end{figure}

Configuration of the model-generation and training processes are controlled via JSONPath-based \cite{jsonpath} rules that annotate the domain schema.  JSONPath is a corollary to XPath \cite{xpath} for XML, but its most-specific-match approach is also similar to CSS \cite{css}, a format familiar to humanists whose primary coding experience is often basic web development.  Rules are applied in order and consist of a JSONPath pattern that matches zero or more locations in the domain schema, and a dictionary of values that will get written at each matching location to a special field, ``meta'', overwriting existing keys.  Figure \ref{figure:jsonpath} shows an example rule that modifies the handling of image properties to down-sample to a particular shape and encoding, and to use \emph{NullDecoder}.  Selective use of null sub-architectures for encoding and decoding allows Starcoder to train models focused on specific tasks: for example, if one categorical property was set to encode with \emph{NullEncoder}, and all other properties to decode with \emph{NullDecoder}, we recover a straightforward classifier for the categorical property.  Starcoder starts with a default list of rules with reasonable defaults.  JSONSchema \cite{jsonschema} templates are used to provide validation of domain schemas, entities, and configuration rules.  This flexibility allows, for example, tight control over parameter counts and how they are allocated across different properties and entity-types.

Ideally, a domain schema is carefully written in the early stages of a humanistic study and adapted as needs dictate, helping to guide the assembly of primary sources, but can often be \emph{automatically derived} to a large degree from common data formats with varying amounts of scholarly guidance.  Starcoder has algorithms for working with several common formats, such as tabular data, Text Encoding Initiative \cite{tei} XML, and SQL.  The ergonomics of this approach can be seen in the large number of studies from the past year spanning a dozen academic departments and listed in Table \ref{table:studies}.

\begin{table}[h!]
  \centering
  \caption{Domain schemas are included in supplementary materials, and the corresponding primary sources will be released on publication.  Entries marked with \(\star\) are datasets previously released in the natural language processing community.}
  \vspace{.4cm}
  \begin{tabular}{lr}
    \toprule
    Title & \multicolumn{1}{r}{Entity count} \\
    \midrule
    \multicolumn{2}{c}{Literary} \\
    \midrule
    Middle English Scansion & 214148 \hspace{.15cm} \\
    Quran & 401 \hspace{.15cm} \\
    Reddit Authorship & 8614 \hspace{.15cm} \\
    Women Writers' Project & 16777 \(\star\) \\
    LitBank & 294411 \(\star\) \\
    \multicolumn{2}{c}{History} \\
    \midrule
    Affiches Americaines & 59422 \hspace{.15cm} \\
    Entertaining America & 14400 \hspace{.15cm} \\
    Indentured Servitude & 14586 \hspace{.15cm} \\
    Caribbean Marronage & 39671 \hspace{.15cm} \\
    Paris Tax Rolls & 621 \hspace{.15cm} \\
    Post-Atlantic Slave Trade & 65163 \hspace{.15cm} \\
    \multicolumn{2}{c}{Ancient Near East} \hspace{.15cm} \\
    \midrule
    Documentary Hypothesis & 24190 \hspace{.15cm} \\
    Cuneiform Digital Library & 92148 \hspace{.15cm} \\
    Royal Inscriptions & 1876 \hspace{.15cm} \\
    \multicolumn{2}{c}{Social Science} \hspace{.15cm} \\
    \midrule
    Ebola Misinformation & 6690 \hspace{.15cm} \\
    Texas Death Row & 1218 \hspace{.15cm} \\
    Vaccine Acceptance & 14099 \hspace{.15cm} \\
    Smoking and Vaping & 124539 \hspace{.15cm} \\
    \multicolumn{2}{c}{Computational Linguistic} \\
    \midrule
    Multimodal Wikipedia & 68424 \hspace{.15cm} \\
    User Fluency & 111378 \hspace{.15cm} \\
    Story Cloze & 476047 \(\star\) \\
    Language ID & 62657 \(\star\) \\
    Sentiment Treebank & 669873 \(\star\) \\
    \multicolumn{2}{c}{Synthetic} \\
    \midrule  
    Arithmetic & 7104 \hspace{.15cm} \\
    \bottomrule
  \end{tabular}
  \label{table:studies}
\end{table}

The remainder of this paper focuses on translating domain schemas into isomorphic models, but we note another critical, practical advantage for collaborating with humanists: a schema allows us to automatically generate extensive web interfaces for jointly introspecting and interacting with data and models.  Such resources, which are sometimes distinguished as ``Public Digital Humanities'' and addressed piecemeal for different studies, would ideally emerge as natural artifacts from the computational, machine learning stage.  We have made significant progress in this direction, and describe the engineering details fully in the appendix along with illustrative screenshots.

\section{Starcoder models}

Starcoder selects and combines appropriate neural sub-architectures to represent the properties, entity-types, and relationship-types defined in a schema, and we now describe this process largely in terms of sub-architecture \emph{function signatures}, with some illustrations from actual studies.  Please see the appendix for complete lists of current implementations and study descriptions, as well as hyper-parameters for the models behind all reported outcomes.

\subsection{Overview}

\begin{figure*}[ht!]
  \centering
  \include{just_starcoder}
  \caption{Model fragment corresponding to the \emph{person} entity-type from Figure \ref{figure:domain} and applied to a person with two clients and an office.}
  \label{figure:starcoder}
\end{figure*}
Figure \ref{figure:starcoder} shows part of a Starcoder model based on employment records, focusing on the structure relevant to a \emph{person} entity-type.\footnote{See appendix for a walkthrough of the corresponding schema and data.}  At a high level, it is an autoencoder: an entity (top left) passes through many transformations, including bottlenecks in red, and is regenerated at the bottom right, with reconstruction error as the loss function.  Internally, the model is isomorphic to the domain definition of a \emph{person}: the \emph{name}, \emph{age}, and \emph{job} properties have corresponding encoders and decoders (green and blue rectangles) responsible for transforming values to and from a fixed-length representation.  An entity's encoded properties are first batch-normalized to help mitigate differences of scale, and passed through an autoencoder to capture patterns involving multiple properties.  Stopping here would actually produce several independent models, one for each entity-type, with no relational awareness (depth \( 0 \)).  To connect these models and allow information to flow between related entities, we stack additional entity-autoencoders up to depth \( D \), where the entity-autoencoder at depth \( d \) takes the output from that at depth \( d - 1 \) concatenated with a transformation of the \emph{bottleneck} representations of related entities for each potential relationship type.  We now discuss each component of the model precisely: 

\subsection{Input components of related entities}
An ideal training instance for a given domain is a connected component of entities from the multigraph defined by all relationship types: in addition to the adjacency matrices describing the relationships, each entity has some subset of properties defined in the schema.  The example shows a component of at least four entities: the person-entity it focuses on, two related person-entities, and one related location-entity.  There may be more entities in the component, e.g. others in the same office, and in many domains the data may constitute a single component.  We have designed several policies for breaking a large component down into reasonable splits and batches.  An important policy for many humanistic studies is to sample components under conditional independence, where a specific set or type of entities is \emph{always} included in each batch, with the rest filled with components sampled from the data \emph{absent} those entities, which are strictly (and often dramatically) smaller.  For example, in the \emph{documentary hypothesis} study we always include the ~70 \emph{book} entities, which allows for a variety of chapters to be randomly chosen, or in the \emph{language identification} study, the linguistic structure of language families creates giant components (e.g. ``Indo-European''), and it makes sense to always include the structure and then sample the conditional components (i.e. individual documents).   Other policies include sampling \emph{snowflakes} (useful for less-hierarchical data like social media) and straightforward entity-level sampling (useful for highly-connected data).

\subsection{Properties}

Each property-type has an associated class that implements a mapping between its human and numeric forms:

\[
  {\textrm{pack}}_p: H_p \rightarrow N_p
  \]
  \[
  {\textrm{unpack}}_p: N_p \rightarrow H_p
\]

For instance, the class for a text property has a lookup from unicode characters to integers with a special \emph{unknown} symbol, which allows for near-bijection between character and integer sequences.  When the human form of a property-type is already numeric the mapping is often identity, or performs minor normalizations like accepting either probabilities or log-probabilities.  This abstraction allows for some interesting flexibility, like using a URL or filename to read in multimedia.  Note that packing and unpacking have no parameters, and only occur when human-readable data is loaded or produced.

\begin{figure}[h!]
  \centering
  \includegraphics[height=.3\textheight]{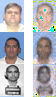}
  \caption{Reconstruction of mug shots from the Texas Death Row test set, a property of type ``image''.}
  \label{figure:mug_shots}
\end{figure}

The numeric form of a property-type may have variable dimensions, such as text or video: since the graph-autoencoding mechanisms accept and produce fixed-length, one-dimensional (FLOD) representations, each property-type requires an \emph{encoder} and \emph{decoder} with signatures:

\[
  {\textrm{encoder}}_p: N_p \rightarrow F_p
  \]
  \[
  {\textrm{decoder}}_p: F_{ae} \rightarrow D_p
\]

where \( F_p \) is the FLOD size of an encoded value of property \( p \), and \( F_{ae} \) is the FLOD size of the output from the final autoencoder in the model.  For example, the default text encoder first embeds the numeric value (a variable-length sequence of integers) then runs it through a GRU, returning the final hidden state.  The default image encoder applies several convolutions and flattens the result.  The decoders are similar but inverse operations.  Figure \ref{figure:mug_shots} shows the input and reconstructed images of mug shots from the Texas Death Row study's test data: while the tiny amount of training (~130 images) is limiting, the model is certainly preserving information through the graph-autoencoding mechanism.

Each property-type also has an associated loss function that is applied to the original numeric value and the \emph{decoded representation}:

\[
{\textrm{loss}}_p: (N_p, D_p) \rightarrow L_p
\]

\subsection{Autoencoding mechanism}
Once an entity's property representations have been computed, they are concatenated into an \emph{entity} representation of length \( R_{e} = \sum \), with zeros corresponding to missing properties.  This is the input to the stack of autoencoders at the center of the model.  While these are currently implemented as vanilla, unnormalized autoencoders \cite{autoencoder}, they need only provide the following signature:

\[
{\textrm{autoencoder}}: R_e \rightarrow (R_e, B_e)
\]

where \( B \) is the size of the bottleneck, or narrowest layer.\footnote{See Section \ref{section:future_work} for plans to constrain the latent space.}  The equal-size input and output allows us to incorporate an actual reconstruction loss from the subarchitecture in addition to that of the overall model, though we have yet to find circumstances where this improves performance.

\subsection{Graph-convolutional mechanism}

Table \ref{table:arithmetic} demonstrates this effect on randomly-generated arithmetic expression-trees, where each node has a categorical \emph{operation} (addition, subtraction, multiplication, division, or constant), relationships (if applicable) to the operation's \emph{left} and \emph{right} arguments, and a calculated scalar \emph{value}.  Training on complete trees and testing with the reported property masked, without graph-awareness the model can only guess mean values (or exploit minor correlations between operation and value), while one-hop graph awareness is a dramatic improvement.\footnote{See appendix for information on hyper-parameters and computational resources of experiments.}

\begin{table}[h!]
  \centering
  \caption{Adding graph-awareness to a model of randomly-generated arithmetic expressions of constants and binary operations with up to 7 nodes goes from random chance to high accuracy.}
  \vspace{.4cm}
  \begin{tabular}{lrr}
    \toprule
    depth & operation (acc) & value (err) \\
    \midrule
    0 & 36.3 & 53.2 \\
    1 & \textbf{94.4} & \textbf{25.0} \\
    \bottomrule
  \end{tabular}
  \label{table:arithmetic}
\end{table}

\subsection{Wiring}
There are many possible variations on how the model is wired and trained, beyond sub-architecture choices: for instance, each autoencoder could employ the directly-encoded properties used at depth \( 0 \), have its own entity-reconstruction loss like at depth \( D \), or explicitly optimize its reconstruction error of the encoded input.  The final decoding process could use the concatenation of all autoencoder outputs (effectively, skip-connections to help with over-parameterization and vanishing gradients).  Efficient batching can require graph-awareness: ideally, connected components would never be split (unless for e.g. dropout), but many structured datasets are a single, large component.  Property encoders can be warm-started independently before use in the full model.  Relationships can be automatically modeled in both directions.  Dropout can be performed on entities, relationships, and properties to encourage (and evaluate) reconstruction of missing information.  The appendix lists the implemented alternatives for each of these sub-architecture and choice.

\subsection{Boosting signal propagation}

The first two columns of Table \ref{table:vanishing} demonstrate graph-depth leading to the perennial issue of vanishing gradients on our \emph{linguisticly-informed language identification}, where training is augmented by the family and genus relationships between languages \cite{wals}, encouraging parameter-sharing.  In this domain, documents from two different, but related, languages will have increasing opportunities to draw from shared representations as depth increases.\footnote{Initially from the ``blank slate'' shared ancestors, and later from those same ancestors' ability to pass through information about other descendants.}  The naive autoencoder stacking from Figure \ref{figure:starcoder} exhibits catastrophic failure at depth 3, as the loss signal fails to traverse the computation graph.  We experiment with two methods to mitigate this problem: first, \emph{highway} connections \cite{highway} such that the output from each autoencoder depth is concatenated for the final decoding and reconstruction process, and \emph{cul-de-sac} losses, where the output from each autoencoder depth is used to attempt reconstruction.  The third and fourth columns demonstrate both methods address the problem, though it is unclear if one is consistently superior.  One practical advantage of the cul-de-sac method is the model can be applied directly with different depths, e.g. for data with less structure or faster run-times, since the intermediary depths have an explicit loss encouraging accurate reconstruction.

\begin{table}[h!]
  \centering
  \caption{Without skip-connections, the model fails to take advantage of structure, and eventually fails catastrophically, while highway connections lead to the best performance at depth 3.}
  \vspace{.4cm}
  \begin{tabular}{lrrr}
    \toprule
    Depth & Naive & Highway & Cul-de-sac \\
    \midrule
    0 & 40.2 & 35.5 & 39.2 \\
    1 & 35.2 & 43.8 & 38.1 \\
    2 & 34.8 & 43.9 & 47.1 \\
    3 &  3.4 & \textbf{49.8} & 37.0 \\
    4 &  3.4 & 43.3 & 40.4 \\
    \bottomrule
  \end{tabular}
  \label{table:vanishing}
\end{table}

\section{The post-Atlantic slave trade in America}

A trained model can serve many purposes, such as distance metrics in retrieval and clustering, reconstruction of missing properties and relationships, interactive probing, and anomaly detection.  To illustrate, consider a recent study of the domestic US slave trade: after the trans-Atlantic trade was banned, there was a large-scale reorganization of the enslaved population, with attendant infrastructure, profit, violence, and disruption to millions of oppressed lives.  The common historical understanding has been that slaves were transported to regions of higher demand, e.g. from Baltimore to New Orleans, by a combination of migrating owners and slave traders, and distributed or sold to nearby sugar plantations or further inland, with little chance of returning to their point of origin.  Tracing or reconstructing the experiences of captives, while critical historically and ethically (e.g. for connecting living descendants to their past), is further hindered by the slave traders' focus on economically salient information, and the asymmetry of recording mechanisms, e.g. mechanism of transport or departure versus arrival port.

\begin{figure*}[h!]
  \centering
  \include{just_manifest}
  \caption{Example manifest from the slave trade between Baltimore and New Orleans, and its transcription into a spreadsheet by the historian.}
  \label{figure:record}
\end{figure*}

To study this process, historians transcribed \cite{williams} ship departure records from the mid-Atlantic region\footnote{Other regions and arrival records are less accessible, or nonexistent for ocean-based trade.} into tabular format, as shown in Figure \ref{figure:record}, leading essentially to a semi-structured, unnormalized database.  We began our collaboration by defining a domain schema with entities like \emph{slave}, \emph{owner}, \emph{ship}, \emph{manifest}, each with appropriate properties and relationships.\footnote{In most cases, intuitive relationships like between \emph{slave} and \emph{owner} entities are indirectly observed, mediated through \emph{manifest} entities.}  Starcoder combined the domain schema with the tabular data to produce a validated dataset of approx. 300k entities, and then generated and trained an associated model of depth 4.  Simply inspecting lists of entity-pairs with highest cosine similarity between bottlenecks highlights the patterns shown in Table \ref{table:most_similar}.

\begin{table}[h!]
  \centering
  \caption{Selected properties of most-similar pairs of a given entity-type.}
  \vspace{.4cm}
  \begin{tabular}{rcl}
    \multicolumn{3}{c}{\underline{\textbf{Increasingly semantic name similarity}}} \\
    William Wiliams & $\Longleftrightarrow$ & William Williams \\
    Baltiomre & $\Longleftrightarrow$ & Baltimore \\
    \multicolumn{3}{c}{\ldots many minor misspellings} \\
    George Y. Kelso & $\Leftrightarrow$ & Kelso \& Ferguson \\    
    New Orleans & $\Leftrightarrow$ & Louisiana \\
    & & \\
    \multicolumn{3}{c}{\underline{\textbf{Slaves sent multiple times from the same port}}} \\
    Louisa, F, 16yo & $\Leftrightarrow$ & Louisa, F, 17yo \\
    Waters, F, 14yo & $\Leftrightarrow$ & Waters, F, 15yo \\
    Kesiah, F, 20yo & $\Leftrightarrow$ & Kesiah, F, 22yo \\
    Taylor, F, 15yo & $\Leftrightarrow$ & Taylor, F, 16yo \\
    \multicolumn{3}{c}{\ldots many more diverse pairings} \\
  \end{tabular}
  \label{table:most_similar}
\end{table}

The first set of examples shows \emph{name} properties from the most-similar entities: the hundreds of typos are actually a major problem for humanists, who may calculate aggregate information with simple spreadsheet queries, but these could also be readily found via string edit distance.  Things then become more interesting, with ``New Orleans'' and ``Louisiana'' having no significant overlap, but are recognized by Starcoder to encode near-identical information.

The second set of examples holds deeper historical interest: the most-similar slave entities are very likely of the same people, considering the dates on the related manifest entities advance along with the age differences.  While the mechanism is unobserved in the records, it appears fairly common for slaves to find themselves returning, sometimes ten or more times, to the same point of departure, only to be reshipped southward.  Our collaborators believe this is evidence of \emph{leasing}, which was perhaps more widespread than previously believed.  This in turn could change the assumptions behind other calculations, such as mortality rates, where a slave disappearing from plantation records has usually been interpreted as death.  Even more interesting is that re-traded young women constitute the top dozen or so most-similar pairings, when overall they form a minority of re-traded slaves.  There are a number of possibilities for why their similarity might be stronger under the model, such as tighter trade-networks or geographic consistency, but the correlations involved are highlighting a particularly grim subset of sexually-exploitative leasing known euphemistically as the ``fancy trade'' \cite{fancy}.

\section{Prior work}
The combination of domain schema and data create a task similar to \emph{representation learning} over \emph{knowledge bases} \cite{kb}.  Graph-convolutional \cite{gcn} and other graph-aware architectures \cite{sage} are an active research area, while the simplest form of autoencoders \cite{autoencoder} continues to be a fundamental concept in more sophisticated models \cite{aegeoA,aegeoB}.  Starcoder's default sub-architectures are standard in the literature (MLP, CNN \cite{cnn}, GRU \cite{gru}), and skip-connections have been found useful for a variety of domains and modalities \cite{skipA,skipB}.  To our knowledge the particular combination of graph-convolutions and autoencoders via bottleneck representations is novel.

\section{Ongoing and future work}
\label{section:future_work}

The most immediate research directions we are pursuing are as follows:

\textbf{Regularization of the latent representation space} beyond naive autoencoding: variational autoencoders \cite{vae} and normalizing flows \cite{normalizing} allow prior constraints that will greatly improve generative performance.  A single sample from the multivariate normal could be reused across all autoencoder depths for a given entity, though it would either constrain the bottlenecks to the same size, or require each depth to have a projector to adapt the sample accordingly: it will be interesting to see whether the former option encourages the different depths to learn commensurate representations.

Adaptive and explicit approaches to \textbf{combining loss functions across modalities}: unlike the combined input to the autoencoder mechanism, where normalization techniques can equalize features from different modalities to the same order of magnitude that the downstream architecture can \emph{then} learn more robust transformations of, deterministic loss functions don't have the luxury of post-hoc rebalancing (how many misplaced pixels is a misspelled word worth?).  Recent work has found random coefficient sampling \cite{lossconditional} to be an effective method, and it may also be useful for humanists to directly specify the relative weighting of different forms of evidence.  Similar concerns drive an interest in \textbf{pretraining and warm-start techniques}: in addition to difficulties introduced by heterogeneous properties, small corpora in the humanities present challenges for mainstream methods whose success relies on finding distributional patterns in big data (e.g. word embeddings, image feature extraction).  Starcoder is designed to take advantage of pretrained components, though it will be important to consider how this could introduce bias and circular reasoning in the target domain.  We have begun work on warm-start methods, where each property encoder-decoder pair is initially trained independently to find a reasonable initial state before introducing the graph-autoencoder mechanism that mixes properties and entity-types.

\textbf{Detecting missing or spurious relationships}: classifiers could be introduced as part of the primary training process \cite{relationshipclassification}, even modifying relationships dynamically.  A more modest starting point is to train classifiers after the fact, predicting relationships based on bottleneck representations.  In either case, care must be taken to ensure that information isn't leaked (e.g. don't try classifying a first-order relationship after depth 0), and particular domains may require different approaches to relationship-dropout to create a useful mixture of positive and negative examples.  The addition of position-aware global joint decompositions of adjacency matrices \cite{globalA,globalB} could also provide information not captured by the finite-depth GCN mechanisms or weakened by missing or incorrect relationships.

\textbf{Expanded sub-architectures:}

A number of Starcoder's basic property types and mechanisms rely on recurrent neural nets and simple reductions such as mean and sum operations.  We are experimenting with attentional mechanisms \cite{attention}, both for property encoders/decoders currently implemented as RNNs and for weighted skip-connections within the autoencoder stack, and deep averaging networks \cite{dan} to allow the learning of arbitrary set functions over related-entity bottlenecks.

Please see the appendix for future work on engineering goals related to infrastructure, public humanities, and research ergonomics.  One engineering-centric goal we \emph{will} mention here is a planned implementation of Starcoder using the HaskTorch \cite{hasktorch} framework.  The ability to express type-level constraints on tensor shapes that must unify at compile-time, coupled with Haskell's powerful abstractions to avoid redundancy while also guaranteeing correctness, could be a tremendous benefit for the complex composition of sub-architectures.  Functional dependencies allow relationships between shapes to be specified at the type level, as in this simple declaration that a graph autoencoder's output shape is fully determined by its depth, and input and representation shapes:
\vspace{.2cm}
\lstset{basicstyle=\footnotesize}
\begin{lstlisting}
class GraphAutoEncoder d i b o | d i b -> o
\end{lstlisting}

Note that these are \emph{type-level literals}, meaning they take on concrete values at runtime, but the Haskell \emph{compiler} assures that the composition of sub-architectures remains consistent as the shapes are propagated (and inferred).  HaskTorch's shared backend and inter-operability with PyTorch \cite{pytorch} has recently made it possible to consider the relative strengths and weaknesses of model design in languages emphasizing compiled precision versus runtime flexibility.

\section{Conclusion}
We have presented \emph{Starcoder}, a framework designed to bring modern neural machine learning to bear on the humanities without disrupting the research of computational or traditional scholars by composing architectural components to produce a model isomorphic to the specification of a scholarly domain.  We demonstrated the effectiveness of a novel combination of graph-convolutional and autoencoding mechanisms for domain-specific relationships, and compared methods for preserving signal under increased depth.  Starcoder's potential for interfacing with a broad range of traditional scholarship is reflected in the large number of current studies, whose domain schemas and data are included in the supplementary material.  Finally, an early practical application of Starcoder to records of the post-Atlantic US slave trade are shown to have already yielded valuable historical insights into the obscured experiences of a marginalized people.

\newpage
\bibliography{starcoder}
\bibliographystyle{icml2020}

\newpage
\appendix
\onecolumn
\section{Architectural and training details}

\begin{table}
  \centering
  \begin{tabular}{lrrr}
    \hline
    Property type & Encoder & Decoder & Loss \\
    \hline
    scalar & MLP & MLP & MSE \\
    categorical & Embed & MLP & KLD \\
    text & EmbedGRU & GRU & KLD \\
    distribution & MLP & MLP & MSE \\
    date & MLP & MLP & MSE \\
    place & MLP & MLP & MSE \\
    image & CNN & CNN & MSE \\
    \hline
  \end{tabular}
  \caption{Default implementations for selected property-types.}
  \label{table:property_types}
\end{table}

All studies employ Adam \cite{adam} up to a maximum of \( 200 \) epochs, with initial learning rate \( 0.001 \), patience of \( 10 \), and early stop of \( 20 \).  Table \ref{table:property_types} lists the default sub-architectures for each property-type: embeddings are size \( 32 \), all hidden sizes are \( 128 \), and entity autoencoders were of shape \( (128, 64) \) with a depth of \( 1 \), except where otherwise specified.  All training was on single NVidia 1080tis, and took less then 2 hours.






\section{Engineering details}

\subsection{Implementation language}
Starcoder is currently implemented in PyTorch.  Due to the complex, open-ended space of potential models, constraining and reasoning over function signatures is an increasing challenge: Python's solution via type hints \cite{mypy} is cumbersome and completely lacks active metaprogramming.  We are therefore considering a transition to Haskell and its HaskTorch framework \cite{hasktorch}.  Two of the most important advantages of Haskell's expressive type system are type-level shape constraints on sub-architectures, which provide compile-time guarantees on their composition, and concise abstractions that obviate error-prone redundancies of Python's reliance on duck-typing.

\section{Public and scholar-facing system}
\begin{figure}[h]
  \centering
  \def\nodewidth{0.15}
  \def\nodespace{2}
  \def\nodehspace{4}
  \def\nodevspace{3}
  \begin{tikzpicture}[yscale=-1]
    \node (elasticsearch) at (\eval{0*\nodehspace}, \eval{0*\nodevspace}) {\includegraphics[width=.25\textwidth]{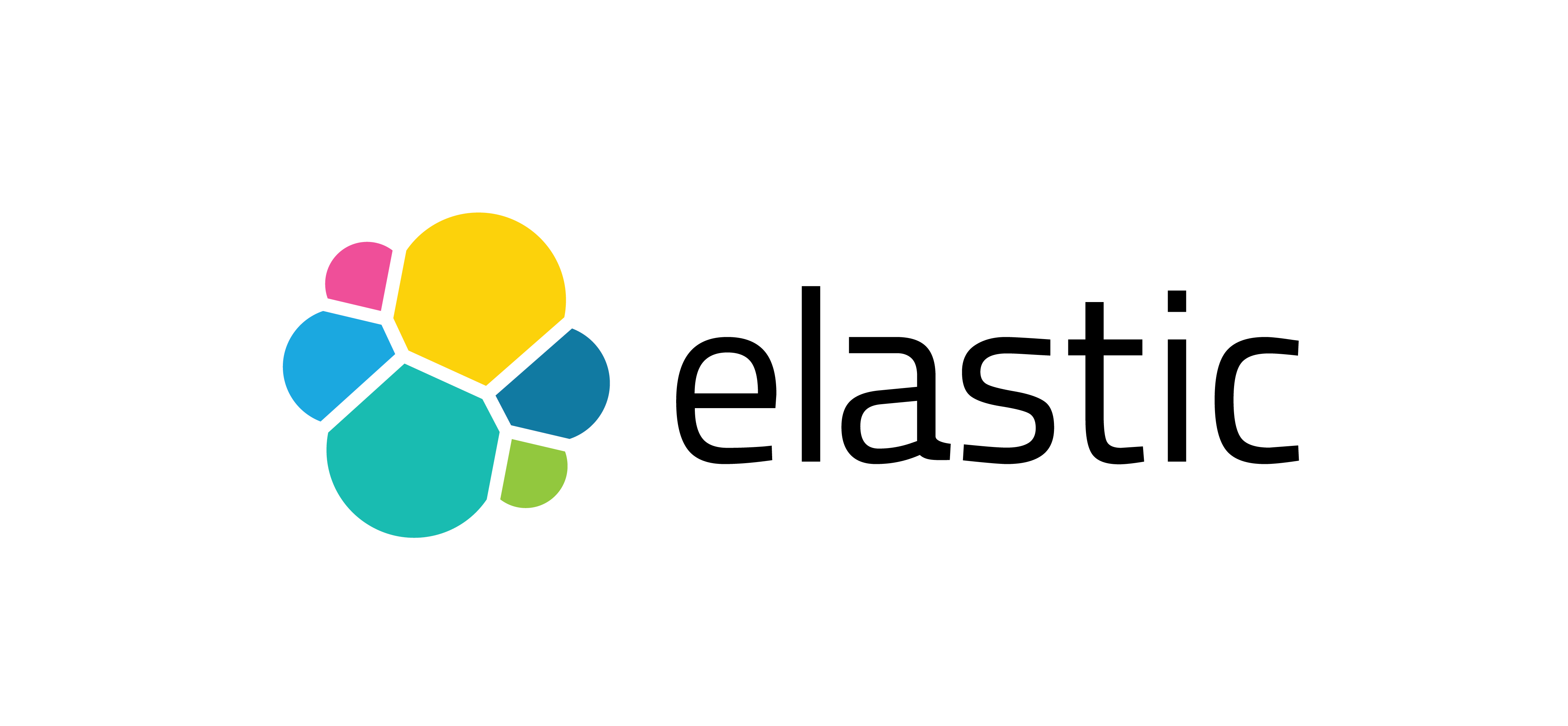}};
    \node (psql) at (\eval{1 * \nodehspace}, \eval{0*\nodevspace}) {\includegraphics[width=.25\textwidth]{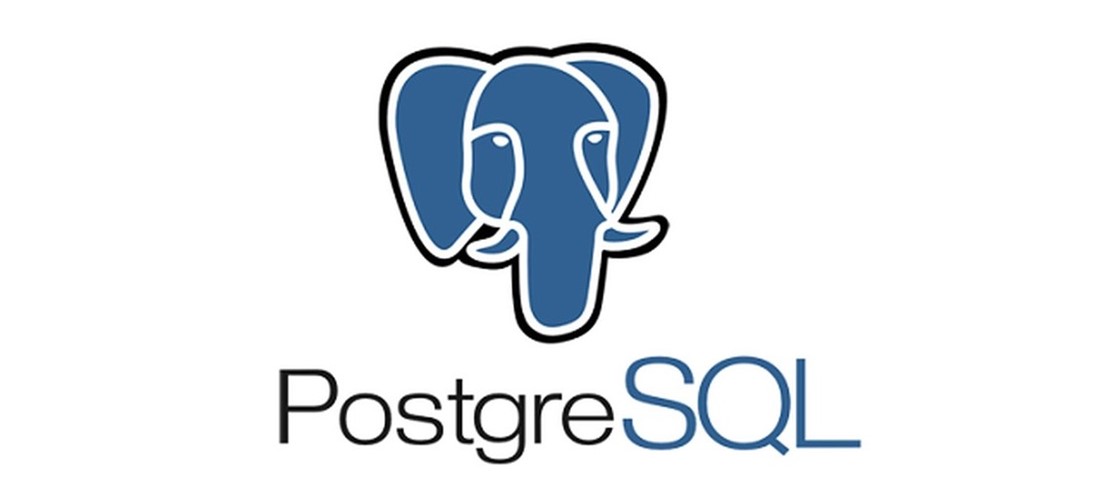}};
    \node (model) at (\eval{2*\nodehspace}, \eval{0*\nodevspace}) {\includegraphics[width=\nodewidth\textwidth]{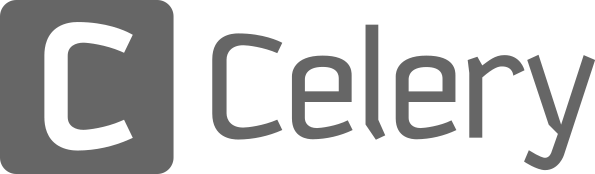}};
    \node (kibana) at (\eval{0*\nodehspace}, \eval{1*\nodevspace}) {\includegraphics[width=\nodewidth\textwidth]{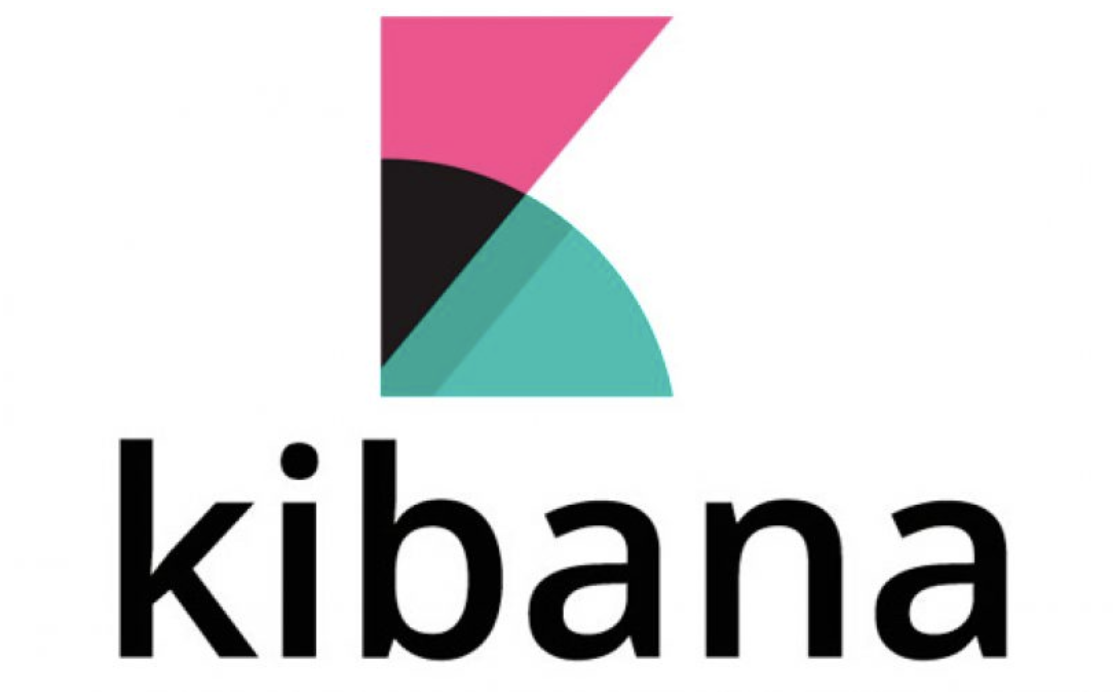}};
    \node (turkle) at (\eval{1*\nodehspace}, \eval{1*\nodevspace}) {\includegraphics[width=.25\textwidth]{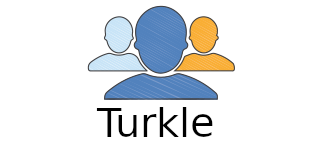}};
    \node (djangogunicorn) at (\eval{2*\nodehspace}, \eval{1*\nodevspace}) {\includegraphics[width=\nodewidth\textwidth]{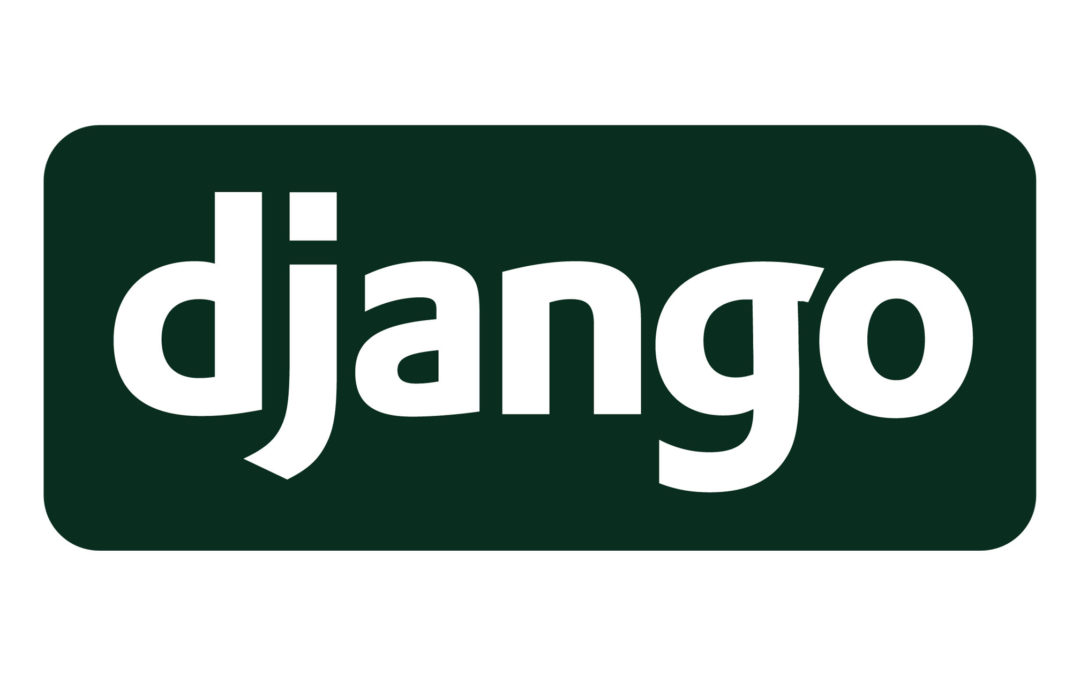}};
    \node (nginx) at (\eval{1*\nodehspace}, \eval{2*\nodevspace}) {\includegraphics[width=\nodewidth\textwidth]{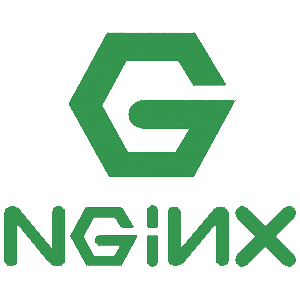}};
    \path (psql) edge[->,bend left=10] (djangogunicorn);
    \path (psql) edge[<-,bend right=10] (djangogunicorn);
    \path (psql) edge[<-] (elasticsearch);
    \path (elasticsearch) edge[->] (kibana);
    \path (djangogunicorn) edge[->,bend left=10] (model);
    \path (djangogunicorn) edge[<-,bend right=10] (model);
    \path (psql) edge[->,bend right=10] (turkle);
    \path (psql) edge[<-,bend left=10] (turkle);
  \end{tikzpicture}
  \caption{The coordinated servers that allow traditional scholars to create projects, explore and annotate their structured data, interact with machine learning models, visualize relationships}
  \label{figure:servers}
\end{figure}
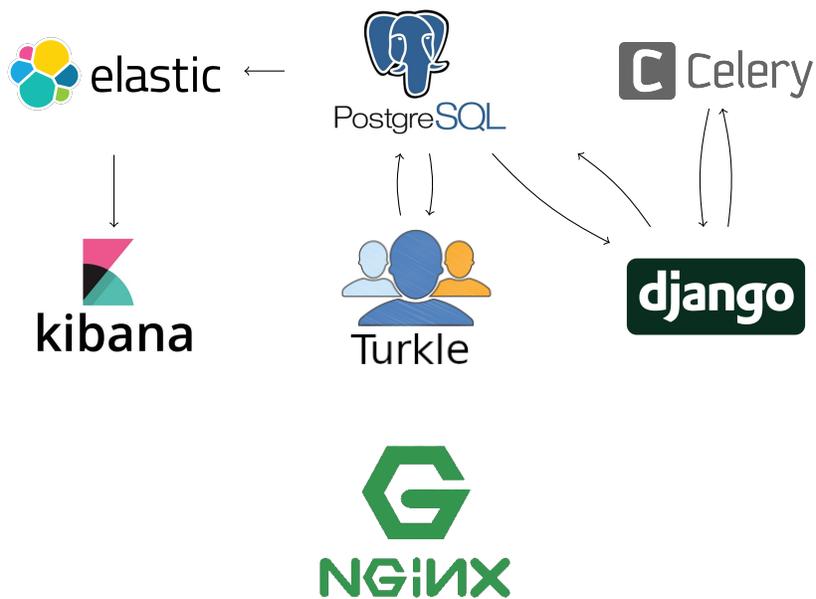

Based on the domain schema, Starcoder can automatically generate a suite of interfaces for interacting with the primary sources and machine learning artifacts.  This includes a ``ground truth'' relational database for the sources, a feature-complete clone of the Mechanical Turk \cite{turk} annotation system, entity browser, clustering based on model bottlenecks, and pair-wise visualizations for properties within and across related entities.  Fast indexing and a subset of visualizations are provided by Elasticsearch and Kibana, while the Django-based core uses Celery to train and apply models on demand.  This allows for a particularly useful feature: editing entities and relationships and rerunning a trained model to dynamically see how properties interact.  A growing suite of web-based editors help humanists to define and enrich domain schemas and experimental settings.  Figure \ref{figure:servers} shows the relationship between these components, and an instance, generated directly from domain schemas and primary sources, can be accessed at \url{https://www.comp-int-hum.org}.

\section{Examples from public and scholar-facing interface}

\begin{figure}
  \centering
  \includegraphics[width=.5\textwidth]{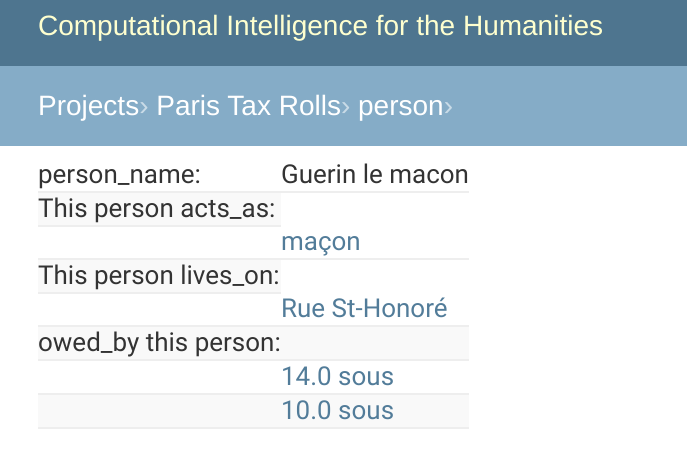}
  \caption{Browser}
  \label{figure:browser}
\end{figure}

\begin{figure}
  \centering
  \includegraphics[width=\textwidth]{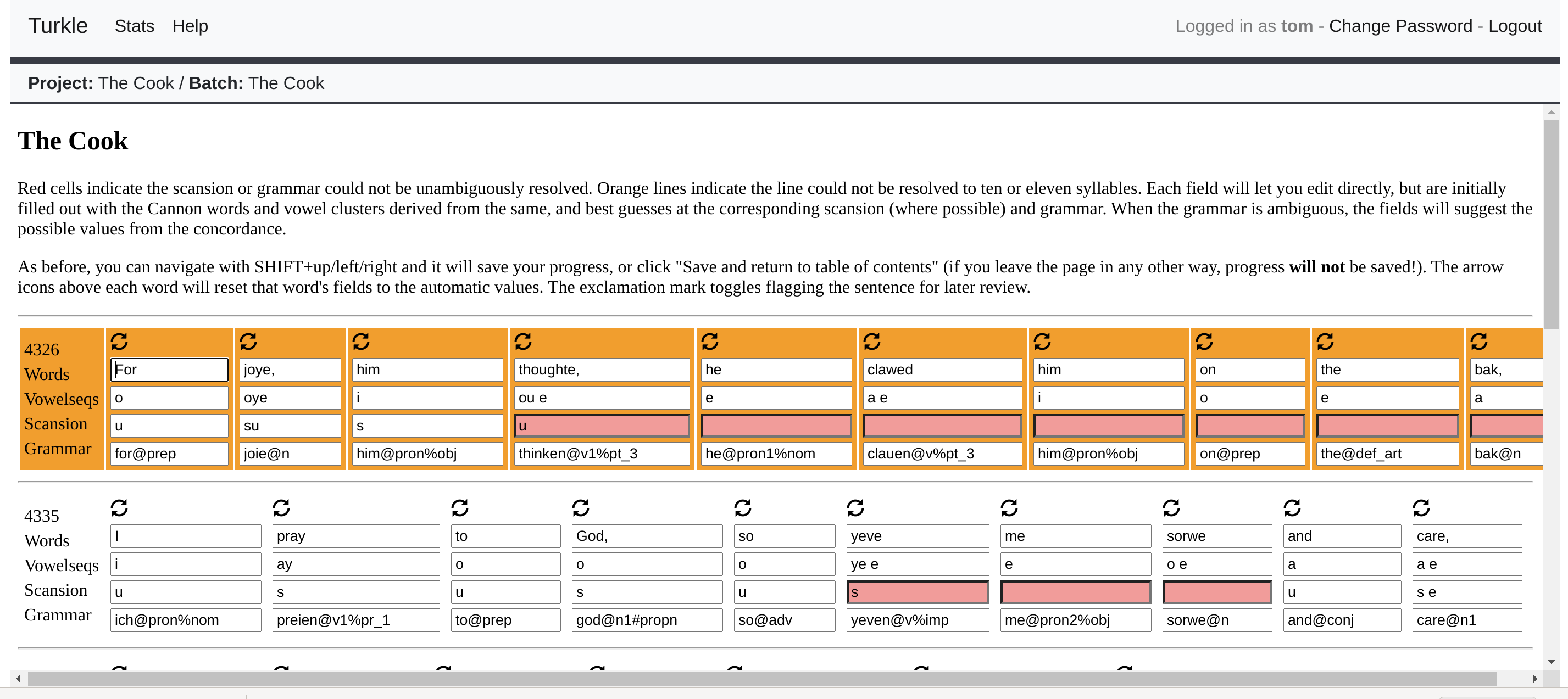}
  \caption{Turkle}
\end{figure}

\begin{figure}
  \centering
  \includegraphics[width=.5\textwidth]{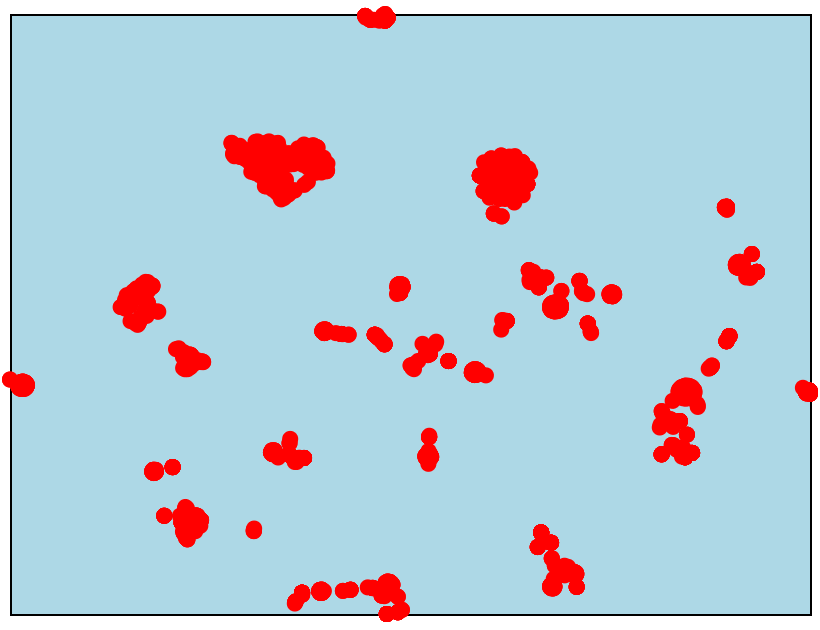}
  \caption{Bottlenecks}
\end{figure}

\begin{figure}
  \centering
  \includegraphics[width=.5\textwidth]{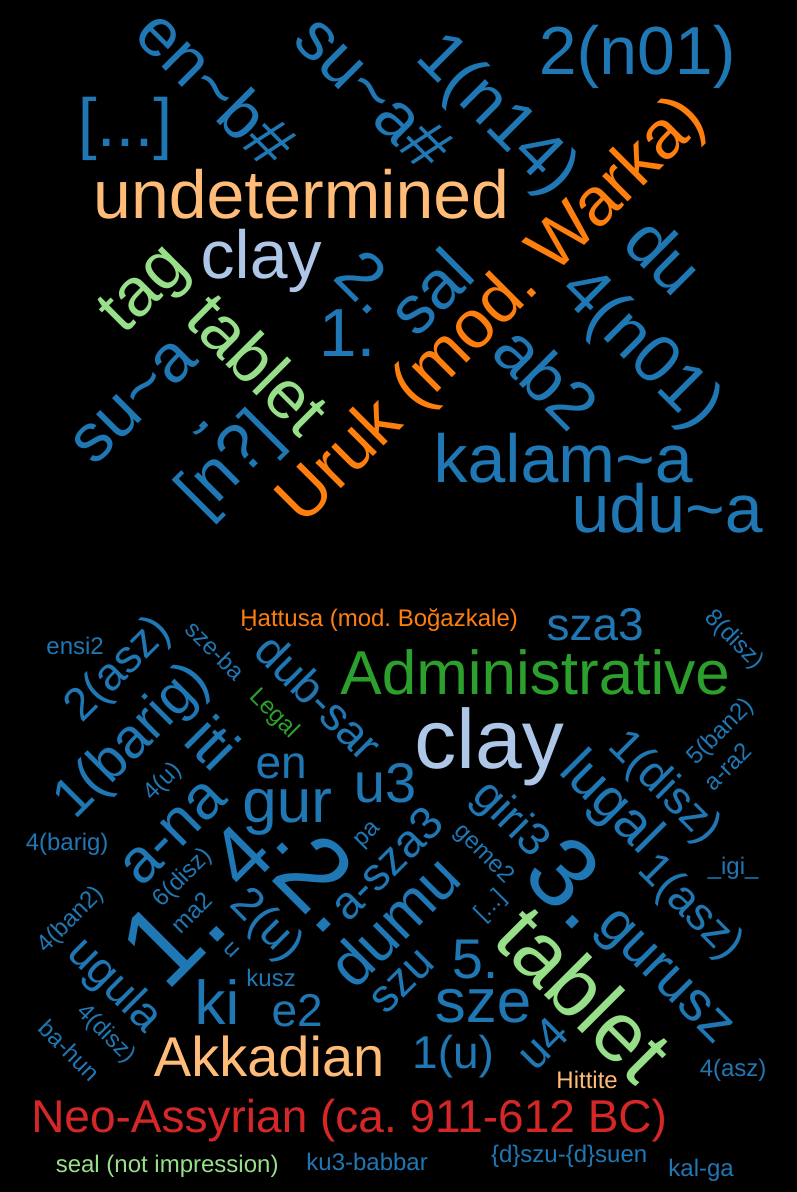}
  \caption{Topics}
\end{figure}

\begin{figure}
  \centering
  \includegraphics[width=.5\textwidth]{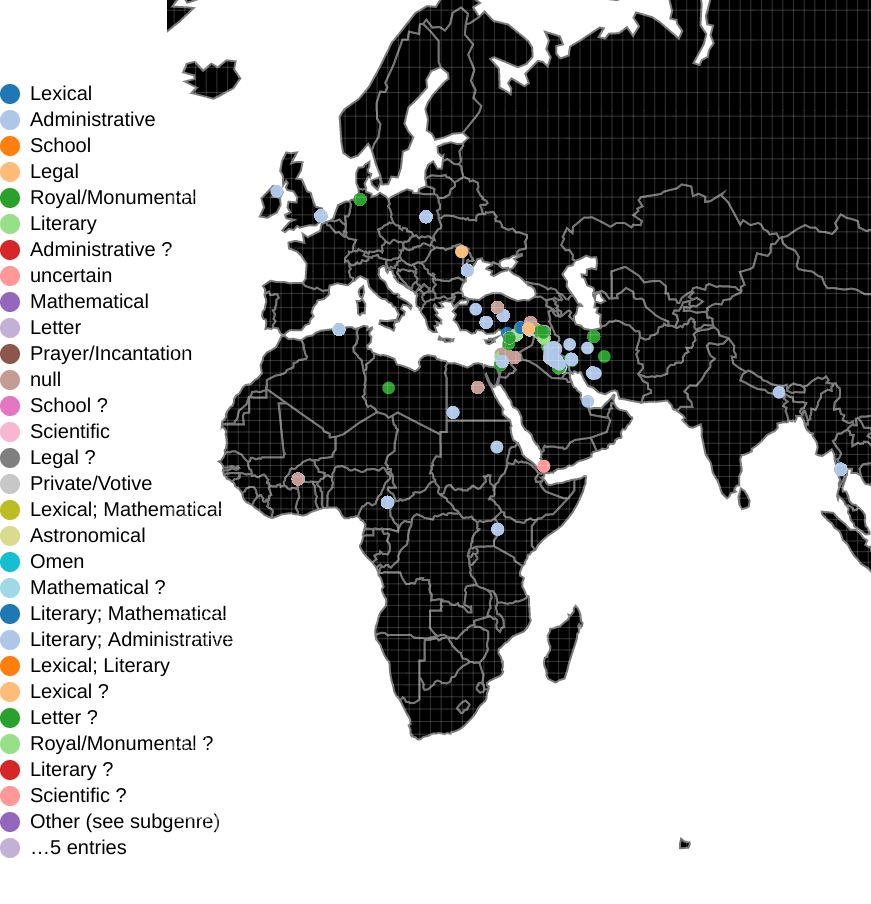}
  \caption{Geographic plot of known and inferred coordinates from the \emph{Cuneiform Digital Library} study.}
\end{figure}

\end{document}